\documentclass[11pt,a4paper]{article}
\usepackage[hyperref]{ranlp2021}
\usepackage{times}
\usepackage{latexsym}

\usepackage{microtype}

\usepackage{pgf-pie}

\aclfinalcopy 

\usepackage[utf8]{inputenc}
\usepackage{graphicx}

\title{Metric Learning in Multilingual Sentence Similarity Measurement for Document Alignment}

\author{Charith Rajitha, Lakmali Piyarathne, Dilan Sachintha, Surangika Ranathunga \\
 Department of Computer Science and Engineering,\\
  University of Moratuwa, Katubedda 10400, Sri Lanka. \\
\texttt{[rajitha.16,lakmali.16,dilansachintha.16,surangika]@cse.mrt.ac.lk}

}

\date{}

\begin{document}
\maketitle
\begin{abstract}
Document alignment techniques based on multilingual sentence representations have recently shown state of the art results. However, these techniques rely on unsupervised distance measurement techniques, which cannot be fined-tuned to the task at hand. In this paper, instead of these unsupervised distance measurement techniques, we employ Metric Learning to derive task-specific distance measurements. These measurements are supervised, meaning that the distance measurement metric is trained using a parallel dataset. Using a dataset belonging to English, Sinhala, and Tamil, which belong to three different language families, we show that these task-specific supervised distance learning metrics outperform their unsupervised counterparts, for document alignment. 
\end{abstract}

\section{Introduction}
Document  alignment is an important precursor to build parallel corpora from noisy data sources. Document alignment is also useful in multilingual Information Retrieval, as well as for tasks such as false news detection across different languages. Traditionally, document alignment has been achieved by metadata-based methods ~\cite{Resnik1998,Resnik1999} and translation based methods ~\cite{uszkoreit2010large,Dara2016}. Metadata-based methods rely on exploiting the meta information of the selected data sources, which may be inconsistent across different sources.  On the other hand, translation based methods assume the availability of a Machine Translation (MT) system.    

To overcome these issues, recent research has exploited the use of multilingual sentence representations (multilingual sentence embeddings) such as LASER~\cite{artetxe2019massively}\footnote{https://github.com/facebookresearch/LASER}.
Here, vector representations are derived for documents in both source and target languages. Then, for a given document in the source side, the most similar counterpart is identified from the target side. Euclidean distance and cosine distance are used in existing document alignment systems \cite{uszkoreit2010large,el2020massively}.  However, these similarity metrics cannot be fine-tuned for the selected task or data at hand.\\
The alternative is to use Metric Learning, which focuses on constructing a problem-specific distance metric automatically from data ~\cite{metric-learn}. Metric Learning-based distance measurement techniques have been successfully employed in image classification and image identification tasks \cite{distance-learning-volcano}.\\
In this paper, we apply two such Metric Learning algorithms on multilingual sentence representations to identify similar document pairs in a bilingual setting. More specifically, we experimented with Sparse High-Dimensional Metric Learning (SDML)~\cite{SDML}, and Information Theoretic Metric Learning (ITML) ~\cite{davis-et-al-icml-2007} algorithms. To the best of our knowledge, this is the first work to exploit the use of Metric Learning with respect to multilingual sentence representations for the purpose of document alignment.\\
~\citet{el2020massively}'s system was used as the baseline. This research can be considered as the best in this line of research at the moment. They calculated the distance between each pair of sentence representations of the source and target documents, and took the sum of the calculated distances for sentence pairs with a weighting value to calculate the distance between two documents.  Euclidean similarity was used as the sentence measurement. Sentence embeddings were derived using the LASER toolkit.\\
We used a dataset that contains documents crawled from news websites belonging to three languages Sinhala (Si), Tamil (Ta), and English (En), which belong to three different language families (Indo-Aryan, Dravidian, and Indo-European, respectively)~\cite{sachintha2021exploiting}. Note that Sinhala is a low-resource language. In the recent language categorization by~\citet{joshi-etal-2020-state}, Sinhala belongs to class 0, meaning that it has exceptionally limited resources. Tamil is considered as a medium-resourced language. It is categorised as 3, meaning that it has a strong web presence, and a cultural community that backs it.\\
We separately report results for the three language pairs, En-Si , En-Ta and Si-Ta. For all the document alignment tasks except one news dataset in En-Si, Metric Learning based distance measurement performed better than the unsupervised distance measurement techniques.  To train the Metric Learning models, a very small parallel corpus of 5000 words were used. Further experiments showed that the content (i.e.~the domain) in the parallel corpus has no impact on the performance of Metric Learning models. The above results were obtained using LASER embeddings, to be a fair comparison with \citet{el2020massively}.  We  experimented with XLM-R multilingual embeddings~\cite{conneau2019unsupervised}\footnote{multilingual BERT (mBERT) ~\cite{devlin2018bert} was not used since Sinhala is not included in mBERT.} as well, using ITML Metric Learning model. Since XML-R showed superior performance over LASER, we conducted extensive experiments where the ITML model was built using different amounts of parallel data with respect to XLM-R. Our results show that the ITML model trained with XLM-R embeddings works equally well even with 1000 parallel sentences. Our code is publicly released\footnote{\url{https://github.com/dilanSachi/kishkyImplementation}}.\\
The rest of the paper is organized as follows. Section 2 discusses related work and Section 2 discusses the basics of Metric Learning. Section 4 presents our methodology. Section 5 provides an overview of the datasets used for the experiments reported in Section 6. Section 7 reports results and finally Section 8 concludes the paper.

\section{Related Work}
\label{realted_work}
\subsection{Early Document Alignment Systems}
Early work on document alignment mainly relied on metadata based ~\cite{Resnik1998,Resnik1999} and translation based approaches ~\cite{Dara2016}. Even though most metadata based systems are language independent as they only use information such as the HTML structure and sentence and document length, these methods tend to have lower results due to the inconsistency of these metadata across different data sources. Translation based approaches outperformed metadata based systems since they depend on the textual context of the documents ~\cite{Vos2004}. However the accuracy of these highly depends on the used MT system.\\
\subsection{Multilingual Embedding Based Document Alignment Systems}
According to ~\citet{schwenk-douze-2017-learning}, a sentence embeddings is a representation of a sentence that is independent of the language and is likely to capture the underlying semantics. Multilingual sentence embeddings allow us to insert sentences from different languages as vectors into a common high-dimensional, language-agnostic semantic space. These methods represent documents using vector representations, where representations of similar sentences have a lower vector distance. Distance between these sentence representations is used to calculate the distance between documents, and documents that have a lower distance are selected as parallel documents.\\
First notable work on multilingual sentence embeddings based document alignment is by~\citet{Seo2016}, who introduced the Hierarchical Attention Networks (HAN) architecture.~\citet{Guo2019} extended the HAN architecture for parallel document mining.~\citet{Guo2019} compared performance of the HAN architecture for document alignment with a neural Bag of Words (BoW) document embedding model, where document embeddings were generated by simply averaging multilingual sentence embeddings.\\
~\citet{el2020massively} proposed a method that uses the LASER toolkit to create the multilingual sentence embeddings. They calculated the distance between each pair of sentence representations of the source and target documents, and took the sum of the calculated distances for sentence pairs with a weighting value to calculate the distance between two documents. Euclidean distance was used as the distance metric for calculating the distance between sentence embeddings. Greedy movers distance algorithm ~\cite{el2020massively}, which is an improved version of the Earth Movers Distance algorithm ~\cite{earth-movers}, was used to take the sum of distance between sentences. They have used multiple sentence weighting schemes such as sentence length, inverse document frequency (IDF) and sentence length combined with IDF (SLIDF) to improve the results further. However, their dataset is not publicly available.\\
\subsection{Multilingual Contextual Embedding Models}
The LASER model consists of a single encoder implemented using a biLSTM (bi-Long Short Term Memory) network, which can handle multiple languages. This guarantees that sentences that are semantically similar lie closer in the embedding space. This encoder is coupled with an auxiliary decoder, and is pre-trained on 93 languages (using parallel corpora). Sentence embeddings are obtained by applying a max-pooling operation over the output of the encoder and used to initialize the decoder LSTM through a linear transformation. The encoder and decoder are shared by all the languages and for that, a joint byte-pair encoding (BPE) vocabulary made on the concatenation of all training corpora was used.\\
The XLM-R model is based on a Transformer model~\cite{vaswani2017attention}. It has an encoder trained only with a masked language model objective. In essence, XLM-R is the multilingual version of RoBERTa~\cite{liu2019roberta}. RoBERTa improves on the popular BERT model~\cite{devlin2018bert} with more data, larger batch sizes, longer training times, training on longer sequences and dynamically changing the masking pattern applied to the training data. XLM-R is trained with common crawl data from 100 languages, and has shown to outperform mBERT on multiple Natural Language Processing tasks.

\section{Metric Learning}
\label{metric-learning}
Unsupervised metrics such as Euclidean or cosine are commonly used to calculate the distance between two embeddings or vectors. This is the same for sentence embeddings.  However, these unsupervised metrics cannot be optimised for the particular data set or the task. In contrast, Metric Learning uses a supervised algorithm to learn the best distance measurement metrics that are task specific, using the given data. Metric Learning has been successfully used for tasks such as pattern recognition and face identification in the field of image classification ~\cite{distance-learning-volcano,pattern-recog,learning-mahalanbis}.
All the Metric Learning algorithms use Mahalanobis distance as the distance metric. Mahalanobis distance between two points \( x,x^{'} \)  is defined as in Equation ~\ref{mahalanobis-1},
\begin{equation}
\label{mahalanobis-1}
 D_{L}(x,x^{'}) = \sqrt{(Lx-Lx^{'})^{T}(Lx-Lx^{'})}
\end{equation}
This can also be written as shown in Equation ~\ref{mahalanobis-2},
\begin{equation}
\label{mahalanobis-2}
D_{L}(x,x^{'}) = \sqrt{(x-x^{'})^{T}M(x-x^{'})}
\end{equation}
where \( M = L^{T}L\).
The Metric Learning algorithms calculate the matrix \(L\) according to the given training data set. 

Commonly used Metric Learning algorithms include: Neighborhood Components Analysis (NCA) ~\cite{NIPS2004_2566}, Large Margin Nearest Neighbors (LMNN) ~\cite{LMNN}, Sparse High-Dimensional Metric Learning (SDML) ~\cite{SDML}, and Information Theoretic Metric Learning (ITML) ~\cite{davis-et-al-icml-2007}. \\
ITML algorithm is able to learn a distance function that generalizes well to unseen test data and is also fast and scalable.  \citet{davis-et-al-icml-2007} used LogDet divergence for regularization of Mahalonbis distance learning methods. This regularization aims at keeping the learned distance close to the Euclidean distance. It has been shown that ITML converges to a global optimal solution.
SDML algorithm aims on training an accurate distance metric for high dimensional data from a small sample. In addition to the LogDet Divergence regularization introduced by \citet{davis-et-al-icml-2007}, L\textsubscript{1}-regularization on the off diagonal elements of the metrix was used to reduce the chance of over fitting. 

\section{Methodology}

\subsection{Baseline Document Alignment System}
\label{baseline}
The document alignment system proposed by~\citet{El-Kishky2020} was used as the baseline, which was discussed in Section~\ref{realted_work}. In this method, first the sentence representations are derived from a multilingual contextual embedding model. 
The algorithm first calculates the Euclidean distance between each sentence pair from source to target. The calculated weight of the sentence is used as the weights (aka mass) when calculating the Greedy Movers Distance between document pairs. Then the minimum value of the weights is selected as the flow value. Then the distance of the sentence pair is updated by adding the multiplication of flow value and the subtraction of two weight values (s\textsubscript{A}) and (s\textsubscript{B}) as shown in Equation \ref{eq-flow}.
\begin{equation}\label{eq-flow}
    distance = distance + ||s_{A} - s_{B}|| \times flow
\end{equation}
In this method, each document is represented as a normalized bag of sentences. Instead of assigning equal weights to each sentence, they have used three weighting schemes for sentences: (1) sentence length (SL) weighting, (2) Inverse Document Frequency (IDF) weighting and (3) sentence length and IDF (SLIDF) weighting.

\subsubsection{Sentence Length Weighting}
\label{sentence-length-weighting}
In the dataset we used, on average, there are 10 sentences per document. However, there is a great variance in the length of the sentences. Needless to say, longer sentences tend to contain more content than the shorter ones. Therefore, longer sentences should have more consideration when aligning documents. Sentence Length weighting scheme uses the sentence length for consideration when calculating the weights.\\
In this weighting scheme, each sentence is weighted using the ratio between the number of tokens into the length of the sentence and the sum of number of tokens into the length of the sentence for each sentence.
For the i\textsuperscript{th} sentence in document A, weight \(d_{A,i}\) is calculated using the equation \ref{sent-len-weighting}.
\begin{equation}
\label{sent-len-weighting}
d_{A,i} = \frac{count(i)\times|i|}{\sum_{s\subseteq A}^{} count(s)\times|s|}
\end{equation}

where \(|i|,|s|\) - Number of tokens in sentence i and s respectively.

\subsubsection{IDF Weighting}
\label{idf-weighting}
The IDF weighting scheme is a common weighting scheme in Information Retrieval. If some content is occurring a number of times in every document, that could be less semantically informative. That could be the titles, and other website specific text. Using this fact, these sentences should get a less weight than other sentences. IDF Weighting scheme is calculated using the equation \ref{idf-weighting}, 
\begin{equation}
\label{idf-weighting}
 d_{A,i} = 1 + log \frac{N+1}{1 + |d \subseteq D : s \subseteq d |}
\end{equation}
where N - Total number of documents and \newline
\( |d \subseteq D : s \subseteq d | \) - Number of documents in which sentence \(s\) occurs.
\begin{table*}[h!]
\begin{tabular}{|p{2cm}|p{1cm}|p{1cm}|p{1cm}|p{1cm}|p{1cm}|p{1cm}|p{1cm}|p{1cm}|p{1cm}|} 
\hline
\multicolumn{1}{|c|}{} & \multicolumn{3}{c|}{Sinhala - English} & \multicolumn{3}{c|}{Tamil - English} & \multicolumn{3}{c|}{Sinhala - Tamil}\\
\cline{2-10}
\multicolumn{1}{|c|}{Web site} & Sinhala & English & GA & Tamil & English & GA & Sinhala & Tamil & GA \\
\hline
Army News &  2033 & 2081 & 1848 & 1905 & 2081 & 1671 & 2033 & 1905 & 1578\\
Hiru &  3133 & 1634 & 1397 & 2886 & 1634 & 1056 & 3133 & 2886 & 2002 \\
ITN & 4898 & 1942 & 352 & 1521 & 1942 & 112 & 4898 & 1521 & 34\\
News First & 1819 & 2278 & 344 & 2333 & 2278 & 316 & 1819 & 2333 & 97\\
\hline
\end{tabular}
\caption{Document alignment data set with golden alignment counts \newline GA - Golden Alignment}
\label{dataset}
\end{table*}

\subsection{Supervised Distance Metric}
\label{supervised-distance-metric}
As our main contribution, we used a supervised distance metrics calculated Metric Learning algorithms instead of using unsupervised metric used by \citet{el2020massively}. 

\begin{table*}[h!]
\centering
\begin{tabular}{|c|c|c|c|c|c|c|}
\hline
Language & Weighting Scheme & Distance Metric & \multicolumn{4}{|c|}{News Site} \\
\hline
&&&Hiru & ITN & Newsfirst & Army News \\
\hline
En - Si & SL & Euc & 0.82 & 0.85 & 0.88 & 0.99\\
&& Cosine & 0.82 & 0.82 & \textbf{0.91} & 0.99 \\
& & \textit{SDML-OPUS-Subtitle Corpus} & \textit{0.85} & \textit{0.87} & \textit{0.90} & \textit{\textbf{0.99}} \\
& & SDML - NLPC Corpus & 0.84 & 0.87 & 0.89 & 0.99 \\
&& \textit{ITML-OPUS-Subtitle Corpus} & \textit{0.84} & \textit{0.85} & \textit{0.89} & \textit{\textbf{0.99}} \\
& & ITML - NLPC Corpus & 0.83 & 0.85 & 0.88 & 0.99 \\
& IDF & Euc & 0.78 & 0.84 & 0.81 & 0.98 \\
&& Cosine & 0.77 & 0.80 & 0.82 & 0.96 \\
& & \textit{SDML-OPUS-Subtitle Corpus} & \textit{\textbf{0.85}} & \textit{\textbf{0.89}} & \textit{0.87} & \textit{0.99} \\
&& \textit{ITML-OPUS-Subtitle Corpus} & \textit{0.82} & \textit{0.84} & \textit{0.86} & \textit{0.98} \\
\hline
En - Ta & SL & Euc & 0.26 & 0.44 & 0.41 & 0.69\\
&& Cosine & 0.30 & 0.47 & 0.50 & 0.78 \\
& & SDML & 0.41 & 0.66 & 0.62 & 0.89 \\
&& ITML & \textbf{0.48} & \textbf{0.67} & \textbf{0.67} & \textbf{0.91} \\
& IDF & Euc & 0.24 & 0.50 & 0.37 & 0.57 \\
&& Cosine & 0.27 & 0.52 & 0.45 & 0.68 \\
& & SDML & 0.46 & 0.64 & 0.56 & 0.82 \\
&& ITML & 0.43 & 0.66 & 0.59 & 0.84 \\
\hline
Si - Ta & SL & Euc & 0.45 & 0.41 & 0.63 & 0.83\\
&& Cosine & 0.50 & 0.64 & 0.60 & 0.88 \\
& & SDML & \textbf{0.57} & 0.61 & \textbf{0.74} & \textbf{0.91} \\
&& ITML & 0.51 & 0.47 & 0.64 & 0.88 \\
& IDF & Euc & 0.42 & 0.47 & 0.59 & 0.73 \\
&& Cosine & 0.44 & 0.61 & 0.59 & 0.77 \\
& & SDML & 0.53 & \textbf{0.64} & 0.71 & 0.86 \\
&& ITML & 0.47 & 0.61 & 0.60 & 0.80 \\
\hline
\end{tabular}
\caption{Recall values for Document Alignment with LASER embeddings}
\label{eval-result-si-en}
\end{table*}

\begin{table*}[h!]
\centering
\begin{tabular}{|c|c|c|c|c|c|}
\hline
Language & Parallel Corpus Size & \multicolumn{4}{|c|}{News Site} \\
\hline
&&Hiru & ITN & Newsfirst & Army News \\
\hline
En - Si & 1000 & 0.90 & 0.92 & 0.93 & 0.99\\
& 2000 & 0.90 & 0.91 & 0.94 & 0.997 \\
&  3000 & 0.90 & 0.91 &0.94 & 0.998 \\
& 4000 & 0.91 & 0.92 & 0.94 & 0.998 \\
&  5000 & 0.91 & 0.92 & 0.94 & 0.998 \\
\hline
En - Ta  & 1000 & 0.82 & 0.94 & 0.93 & 0.989\\
& 2000 & 0.83 & 0.95 & 0.93 & 0.986 \\
& 3000 & 0.83 & 0.95 & 0.93 & 0.988 \\
& 4000 & 0.82 & 0.95 & 0.92 & 0.988 \\
&  5000 & 0.82 & 0.95 & 0.93 & 0.986 \\
\hline
Si - Ta & 1000 & 0.80 & 1.00 & 0.90 & 0.947\\
& 2000 & 0.80 & 1.00 & 0.90 & 0.949 \\
& 3000 & 0.80 & 1.00 & 0.91 & 0.945 \\
& 4000 & 0.79 & 1.00 & 0.90 & 0.944 \\
&  5000 & 0.79 & 1.00 & 0.90 & 0.942 \\
\hline
\end{tabular}
\caption{Recall values for Document Alignment for XLM-R Embeddings with ITML for different parallel dataset sizes}
\label{eval-result-xlm-r_laser}
\end{table*}
We used both Sparse High-Dimensional Metric Learning (SDML) ~\cite{SDML} and Information Theoretic Metric Learning (ITML) algorithms ~\cite{davis-et-al-icml-2007,metric-learn} for our experiments. To provide a supervised signal to the Metric Learning algorithms, a parallel corpus was used. The sentences in the parallel corpus were converted to embedding pairs. The Metric Learning models were trained using these embeddings. Once the model is trained, it is able to provide the distance value between new sentence embedding pairs. We replaced the unsupervised distance calculation method used in the baseline~\cite{el2020massively} with these trained supervised distance metrics to calculate the distance between documents.

\subsection{Date-wise filtering}
In most of the cases, an article corresponding to a news is published in all languages within the same day. Therefore, using this identified property of news websites, we reduced the search space to a date from the whole web domain. Then the news items from a particular day were selected and aligned. 

\section{Data Set}
\label{dataset}
We used the dataset presented by~\citet{sachintha2021exploiting}. This dataset contains news articles belonging to four news websites published in English, Tamil and Sinhala languages. From each of the websites, around 2000 documents (per language) were selected based on the published date of the articles. Table \ref{dataset} shows the statistics of the dataset used for document alignment. They have identified different characteristics in these articles and have used those to filter out the ground truth alignment. In addition, they have manually aligned the dataset with the help of human annotators. The aligned dataset was veriﬁed by the same annotators by switching the data sets. 

5000 sentence pairs from the parallel corpora published by ~\citet{fernando2020data} were used to train the Metric Learning models. This parallel dataset is specific to the official government documents in Sri Lanka. Thus, in order to verify whether the domain of parallel data has any impact on Metric Learning, a 5000 subset from the Open Subtitle Parallel Corpus \cite{lison-tiedemann-2016-opensubtitles2016} was used for En-Si. This corpus consists of 800k parallel sentences, which are translations of movie subtitles. 

\section{Experiments}
For our experiments we used the python implementation of the two metric learning algorithms~\cite{metric-learn}. \\
The first experiment was to determine whether Metric Learning based distance measurement could outperform the unsupervised counterpart. To be a fair comparison, we used the LASER embedding model as used by \citet{el2020massively}. Euclidean and Cosine unsupervised distance metrics, as well as ITML and SDML Metric Learning based distance metrics were calculated for both SL and IDF weighting schemes. 5000 parallel sentences from ~\citet{fernando2020data}'s parallel corpora were used to train the Metric Learning models for En-Ta and Si-Ta. For En-Si, SDML and ITML were trained with SL weighting for ~\citet{fernando2020data}'s parallel corpus and the open subtitle corpus. IDF was trained with the latter corpus only. This is to measure the impact of the type of content in the parallel corpus on the Metric Learning models.\\
We conducted another experiment to compare the performance of LASER and XLM-R multilingual embedding models. First, the same 5000 sentence pairs from \citet{fernando2020data} were used to obtain XLM-R embeddings. Then ITML was trained with these embeddings. ITML model was selected since it was less computationally expensive. SL weighting scheme was selected since it performed better than IDF on average. 
\\
In order to determine the impact of the parallel dataset size on the performance of Metric Learning models, we trained separate ITML models using 1000, 2000, 3000 and 5000 parallel sentences for all three languages using \citet{fernando2020data}'s corpus. Embeddings were generated for the using XLM-R.
 
\section{Results}
We followed the common practice used for document alignment evaluation in WMT16 document alignment shared task~\cite{buck-koehn-2016-findings}. As these ground truth documents only contain a small fraction of parallel documents, there can be many more valid cross lingual document pairs in these dataset. Therefore we evaluated the aligned document pairs using recall (i.e. what percentage of the aligned document pairs in the golden alignment set are found by the algorithm) from the ground truth data-set, as done by~\citet{Dara2016}. \\
Results of our system against \citet{el2020massively}'s system on LASER embeddings are given in Table~\ref{eval-result-si-en}. It also shows the results of the two parallel datasets used to build the Metric Learning models for En-Si.  How the performance of our system varies with respect to the size of the parallel dataset used with the XLM-R model are given in Table~\ref{eval-result-xlm-r_laser}.\\
Results related to all the distance measurement techniques depend on the news source. If most of the news items are present in all the three languages, this sends a strong signal to the aligner. A very good example is the En-Si pair in Army News. Number of Sinhala and English documents this source is roughly equal.  \\
We can see that the results confirm that Metric Learning based distance measurement has been very effective in aligning documents for all three language pairs. Even for the Newsfirst data source for the En-Si pair, metric learning  (SDML) lags behind cosine similarity only by a very small margin.\\
It is evident that XLM-R significantly outperforms LASER. XLM-R performing very well even for 1000 parallel sentence is very promising - this means that Metric Learning could be employed with languages included in the XLM-R model as long they have a very small parallel corpus. 
Another factor we wanted to investigate is the impact of the level of language representation in the multilingual embedding model on document alignment. Although LASER and XLM-R include data from multiple languages, they do not include data from all the languages in equal amounts. If we hypothesize that the amount of language data is proportional to the language categorization proposed by~\citet{joshi-etal-2020-state}, results related to Sinhala has to be the lowest. However, the results do not convey this message. Thus we believe that factors such as language relatedness and language complexity might be playing a role here. \\

\section{Conclusion}
This paper presented the use of a supervised distance measurement technique to implement multilingual sentence embedding based document similarity calculation. Results show that this supervised approach is being able to consistently outperform unsupervised
similarity measurements for three document alignment tasks. A small parallel corpus is required to train the Metric Learning distance measurements. Thus, these techniques can be employed with respect to low-resource languages as well, as demonstrated with the use of low-resource language Sinhala.\\
The main drawback of Metric Learning algorithms is that they are very much computationally expensive, when trained on large corpora. Therefore, in the future, we plan to optimize these algorithms to achieve better efficiency.  

\section*{Acknowledgments}
This publication was funded by a Senate Research Committee  (SRC) Grant of University of Moratuwa, Sri Lanka.

\bibliographystyle{acl_natbib}
\bibliography{anthology,ranlp2021}


\end{document}